\documentclass[letterpaper, 10 pt, conference]{ieeeconf}  

\IEEEoverridecommandlockouts                              

\usepackage{graphics} 
\usepackage{epsfig} 
\usepackage{mathptmx} 
\usepackage{times} 
\usepackage{amsmath} 
\usepackage{amssymb}  
\usepackage{graphicx}
\usepackage[font=small]{caption}
\usepackage{subcaption}
\usepackage{booktabs}
\usepackage{multirow}
\usepackage{marvosym}
\usepackage{makecell} 
\usepackage{booktabs}
\usepackage{url}
\usepackage{xspace}
\usepackage{siunitx}
\usepackage{hyperref}
\hypersetup{
    colorlinks=true,
    urlcolor=blue
}

\usepackage{censor}

\newcommand{\algname}{SOFTMAP\xspace}

\title{\LARGE \bf
\algname: Sim2Real Soft Robot Forward Modeling via Topological Mesh Alignment and Physics Prior
}

\author{
Ziyong~Ma\textsuperscript{\Letter,1},
Uksang~Yoo\textsuperscript{1},
Jonathan~Francis\textsuperscript{1,2},
Weiming~Zhi\textsuperscript{*,3,4,5},
Jeffrey~Ichnowski\textsuperscript{*,1},
Jean~Oh\textsuperscript{*,1}
\thanks{\textsuperscript{1}Robotics Institute, Carnegie Mellon University, USA;
\textsuperscript{2}Bosch Center for Artificial Intelligence, USA;
\textsuperscript{3}School of Computer Science, The University of Sydney, Australia;
\textsuperscript{4}Australian Centre for Robotics, The University of Sydney, Australia;
\textsuperscript{5}College of Connected Computing, Vanderbilt University, USA;
\textsuperscript{*}Equal Advising.}
}

\begin{document}

\maketitle
\vspace{-20mm}
\thispagestyle{empty}
\pagestyle{empty}



\begin{abstract}
While soft robot manipulators offer compelling advantages over rigid counterparts, including inherent compliance, safe human-robot interaction, and the ability to conform to complex geometries, accurate forward modeling from low-dimensional actuation commands remains an open challenge due to nonlinear material phenomena such as hysteresis and manufacturing variability. We present \algname, a sim-to-real learning framework for real-time 3D forward modeling of tendon-actuated soft finger manipulators. \algname combines four components: (1) As-Rigid-As-Possible (ARAP)-based topological alignment that projects simulated and real point clouds into a shared, topologically consistent vertex space; (2) a lightweight MLP forward model pretrained on simulation data to map servo commands to full 3D finger geometry; (3) a residual correction network trained on a small set of real observations to predict per-vertex displacement fields that compensate for sim-to-real discrepancies; and (4) a closed-form linear actuation calibration layer enabling real-time inference at 30~FPS. We evaluate \algname on both simulated and physical hardware, achieving state-of-the-art shape prediction accuracy with a Chamfer distance of 0.389~mm in simulation and 3.786~mm on hardware, millimeter-level fingertip trajectory tracking across multiple target paths, and a 36.5\% improvement in teleoperation task success over the baseline. Our results show that \algname provides a data-efficient approach for 3D forward modeling and control of soft manipulators. Supplemental materials, experiment data, and visualizations are available at \url{https://ziyongma.github.io/SOFTMAP}.
\end{abstract}



\section{Introduction}


Soft robot manipulators offer compelling advantages over their rigid counterparts, including inherent compliance, safe human-robot interaction, and the ability to conform to complex geometries~\cite{abondance2020dexterous}. The infinite-dimensional deformation space of soft bodies resists compact analytical descriptions, making accurate state representation, estimation, and control challenging~\cite{polygerinos_soft_2017}. Further complicating forward modeling, tendon-driven actuation introduces nonlinear material phenomena such as hysteresis, creep, and manufacturing variability that are difficult to capture from first principles alone~\cite{polygerinos_soft_2017}.



These challenges make accurate real-time 3D forward modeling remain an open problem for soft manipulators. Prior approaches generally fall into two categories, each with its own limitations. Physics-based simulation can generate rich shape data at low cost but suffers from a persistent reality gap, where unmodeled nonlinearities cause predictions to diverge from hardware behavior~\cite{yoo2023toward,yoo2025kinesoft}. Purely data-driven methods can capture real-world dynamics but require prohibitively large annotated datasets collected on physical hardware~\cite{wang2020real}.

\begin{figure}[t]
  \centering
  \includegraphics[width=\columnwidth]{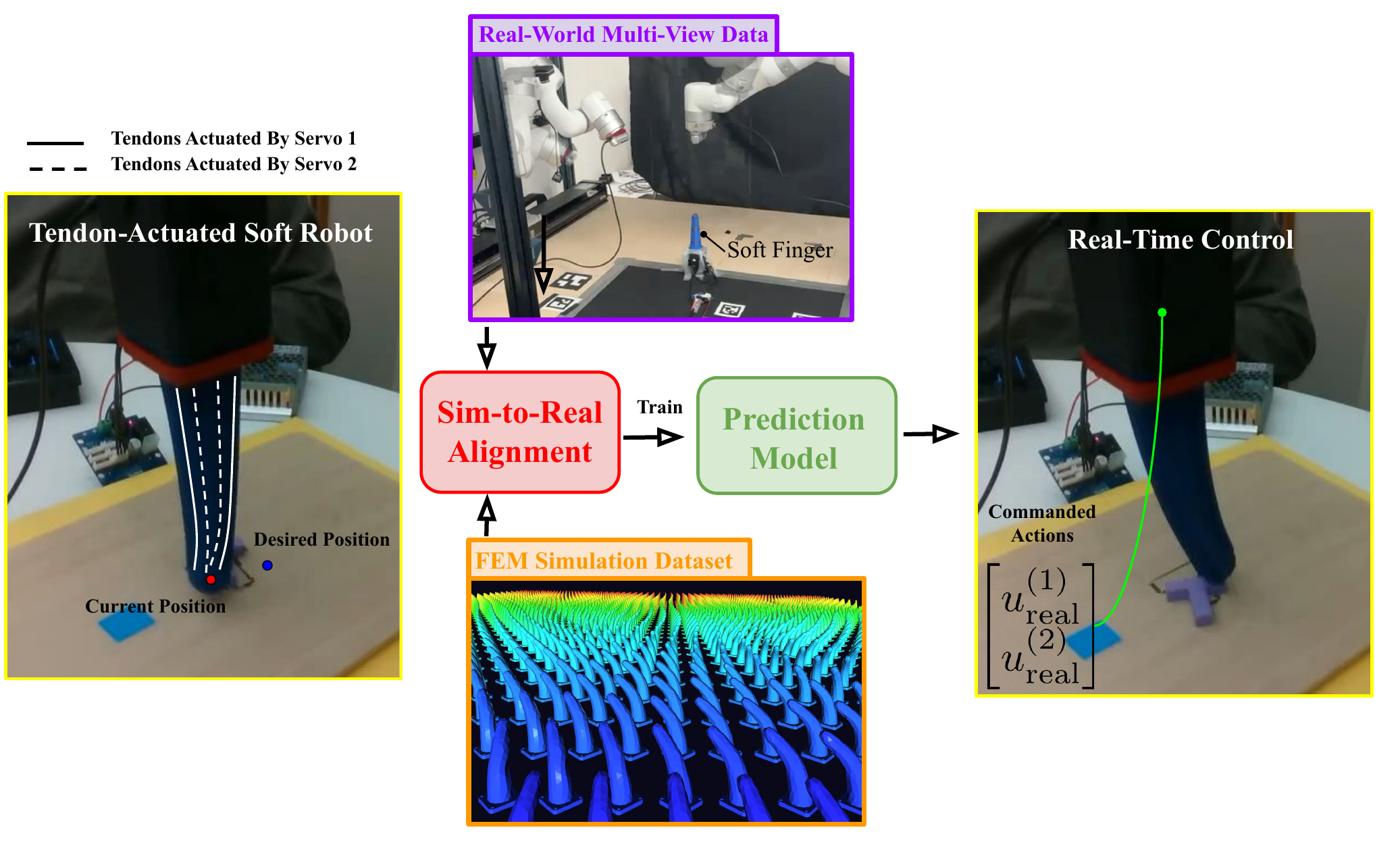} 
  \caption{\textbf{\algname overview.} Simulation and real multi-view data are aligned via ARAP into a shared topology, enabling a lightweight learned forward model to predict full 3D soft finger geometry from servo commands for real-time control and teleoperation.}
  \label{fig:splash}
\end{figure}

In this work, we propose \algname, a sim-to-real learning framework that combines the complementary strengths of both paradigms to achieve accurate, real-time 3D shape prediction of soft finger manipulators. 
The framework consists of four stages. First, we employ As-Rigid-As-Possible (ARAP) deformation~\cite{sorkine2007rigid, yoo2024ropotter} to project both simulated and real point cloud observations into a shared topologically consistent vertex space, establishing reliable cross-domain correspondences that enable accurate sim-to-real transfer with minimal real-world data. Second, we train a lightweight model, \algname's model, on simulation data to serve as a forward model mapping servo commands to full-finger geometry. Third, we train a residual correction network on a small set of real observations to predict per-vertex displacement fields that compensate for residual sim-to-real discrepancies. Finally, a closed-form linear calibration layer aligns the actuation command spaces, enabling real-time inference at high frequencies suitable for closed-loop control.

We evaluate \algname on both simulated and real hardware, and we demonstrate state-of-the-art forward shape prediction accuracy and millimeter-level fingertip trajectory tracking across multiple target paths. We further demonstrate the utility of the learned forward model in downstream applications, including trajectory generation and real-time teleoperation, yielding more accurate fingertip localization and improved teleoperation performance compared to baseline models. The main contributions of this paper are:
\begin{itemize}
    \item We present \algname, a sim-to-real learning framework combining ARAP-based topological alignment, simulation pretraining, and lightweight residual correction for accurate 3D forward modeling of tendon-actuated soft finger manipulators.
    \item A data-efficient residual correction mechanism that bridges the sim-to-real gap by learning per-vertex displacement fields from a small set of real observations, reducing Chamfer distance by 33.4\% over the sim-only model.
    \item Experimental evaluation of \algname on millimeter-level fingertip trajectory tracking and real-time vision-based teleoperation on physical hardware, outperforming baselines across all evaluated tasks.
\end{itemize}

\begin{figure}[t]
  \centering
  \includegraphics[width=\columnwidth]{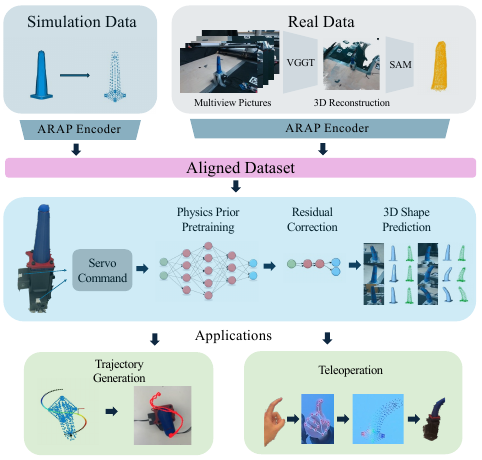}
  \caption{\textbf{\algname pipeline.} Simulation data and real multi-view images are independently processed into 3D representations, then encoded via ARAP into a shared topologically consistent vertex space to form an aligned dataset. A learned MLP maps servo commands to 3D shape predictions, which are subsequently refined by a residual correction network. The resulting forward model enables downstream applications, including trajectory generation and real-time teleoperation.}
  \label{fig:pipeline}
\end{figure}

\section{Related Work}
Accurate forward modeling of soft robot manipulators remains an open challenge, requiring methods that can handle the complexity introduced by soft robots' actuation mechanism, can represent and estimate continuous 3D deformation states, and will enable reliable control despite nonlinear dynamics and sim-to-real gaps.

\subsection{Soft Robot Manipulators}
Soft robot manipulators have garnered significant attention due to their inherent compliance, adaptability, and capacity for safe interaction in unstructured environments~\cite{abondance2020dexterous, yoo2025soft, yao2025soft}. Among actuation strategies, tendon-driven designs~\cite{seleem2025tendon} offer finer control over deformation compared to pneumatic alternatives~\cite{puhlmann2022rbo, yoo2021analytical}, making them well-suited for dexterous manipulation tasks~\cite{bhatt2022surprisingly}. However, the nonlinear material properties inherent to soft bodies, including hyperelasticity, hysteresis, and manufacturing variability, make accurate forward modeling directly from low-dimensional actuation commands fundamentally challenging~\cite{della2023model}. \algname directly addresses this challenge for tendon-actuated soft fingers, learning a forward model that maps servo commands to full 3D finger geometry.

\subsection{Soft Robot State Representation}
Accurate state representation is a prerequisite for model-based control of soft robots. Classical approaches rely on finite element method (FEM) simulations~\cite{qin2024modeling} or Cosserat rod models~\cite{liu2020toward} to approximate the continuous deformation field, but these methods are computationally expensive and sensitive to material parameter uncertainty. Vision-based approaches have emerged as a practical alternative, leveraging RGB images to reconstruct 3D shape~\cite{yoo2023toward} and enable soft robot proprioception through mesh-based and point cloud representations. However, these methods typically operate within a single domain, either simulation or real, limiting their ability to exploit complementary data sources. ARAP deformation~\cite{sorkine2007rigid} provides a principled way to establish topologically consistent shape correspondences across observations, which we leverage to bridge the sim-to-real gap. \algname extends this by projecting both simulated and real point clouds into a shared ARAP-encoded vertex space, enabling cross-domain supervision and accurate forward modeling with minimal real-world data.

\subsection{Soft Robot Control}
Control of soft robots presents unique challenges due to their high degrees of freedom and nonlinear dynamics. Model-based controllers have been developed using analytical forward models~\cite{liu2020toward, della2023model}, but their computational cost and sensitivity to unmodeled nonlinearities limit applicability to real-time settings. Learning-based approaches, including neural network-based forward models, have shown promise in capturing complex input-output mappings directly from data, with DeepSoRo~\cite{wang2020real} demonstrating the feasibility of end-to-end learned models for soft robot shape prediction and control on physical hardware. However, purely data-driven methods require large amounts of real-world data that is costly and time-consuming to collect. Sim-to-real transfer offers a compelling alternative by leveraging cheap simulation data, but the persistent reality gap due to unmodeled material nonlinearities remains a key obstacle~\cite{yoo2023toward}. \algname addresses this by combining a simulation-pretrained forward model with lightweight residual correction to bridge the reality gap, and a model-based inverse kinematics solver that enables accurate real-time trajectory tracking on physical hardware without requiring large real-world datasets.

\begin{figure}[t]
  \centering
  \includegraphics[width=\columnwidth]{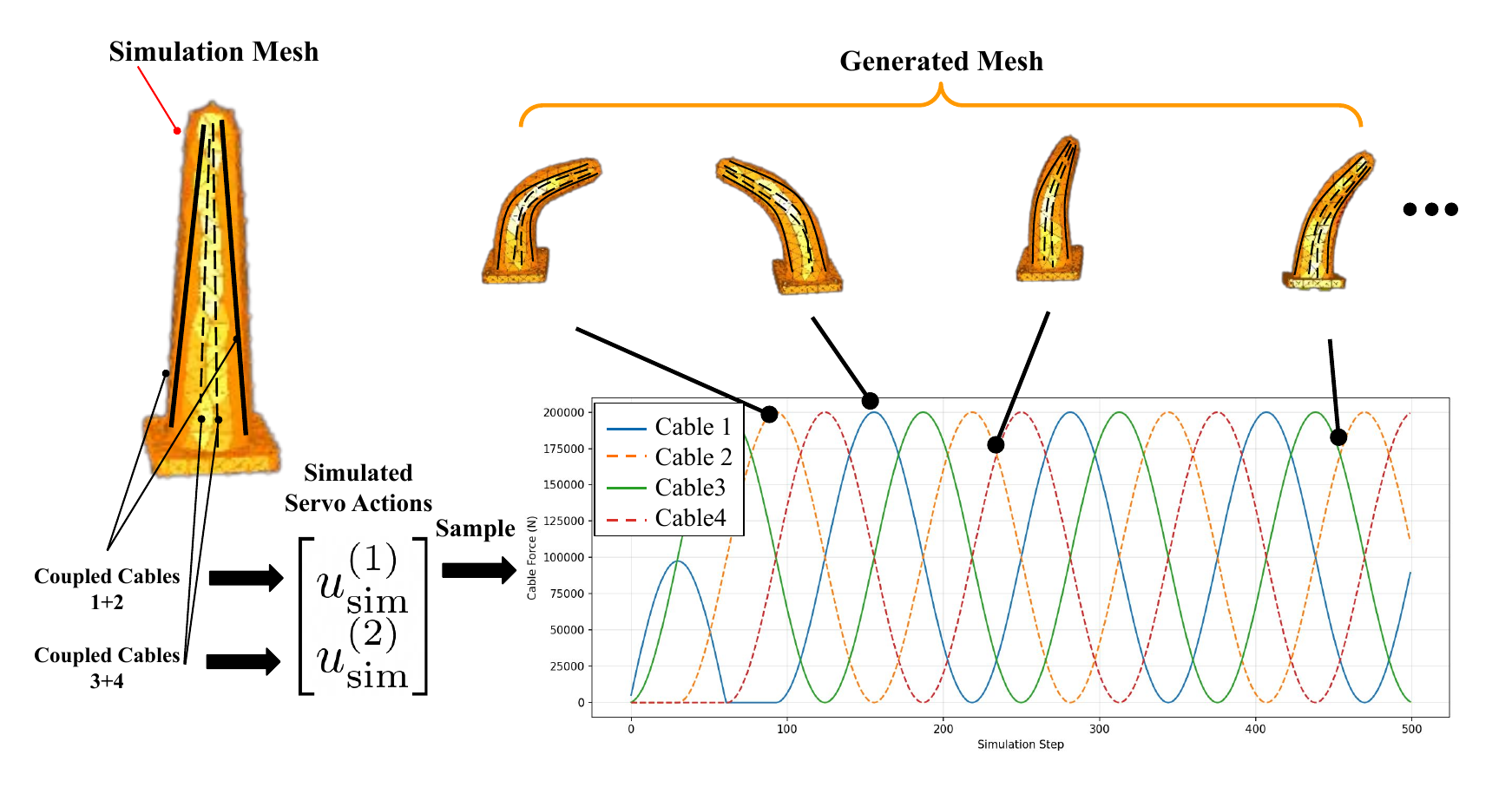}
  \caption{\textbf{Simulation environment.} The soft finger is modeled in the SOFA Framework using a Neo-Hookean hyperelastic material with four embedded tendons. Actuation commands $u_{\text{sim}}^{(i)} \in \mathbb{R}^2$ are swept across a dense grid to generate shape observations $Y_{\text{sim}}^{(i)} \in \mathbb{R}^{548 \times 3}$, forming the simulation dataset used to pretrain \algname's model.}
  \label{fig:sim_setup}
\end{figure}

\section{Method}

\algname aims to achieve real-time, accurate 3D forward modeling for tendon-actuated soft finger manipulators by bridging the reality gap between physics-based simulation and physical hardware. A core challenge is that purely data-driven methods require prohibitively large real-world datasets, while purely simulation-based methods fail to capture unmodeled physical nonlinearities such as material hysteresis, manufacturing variability, and sensor noise. To address this, \algname combines a simulation-pretrained neural network as a physics prior with a lightweight residual correction network, enabling accurate sim-to-real transfer with minimal real-world data. In this work, we use the open-source tendon-driven soft-robot platform MOE~\cite{yoo2025soft}.

\subsection{Data Collection}
To enable supervised sim-to-real learning for the shape forward model, we collect simulation and real-world deformation data of a soft finger.
Simulation is done modeling tendon actuation in SOFA Framework~\cite{westwood2007sofa} and in the real world using two robot arms.

\textbf{Simulation.} To generate clean and physics-consistent data that match the real physical finger, we simulate a soft elastomer finger using a Neo-Hookean hyperelastic material model within the SOFA Framework~\cite{westwood2007sofa}. The finger has four tendons embedded along the finger's length to mimic cable-driven actuation, with tendon tensions serving as the control inputs. As shown in Fig.~\ref{fig:sim_setup}, to cover the finger's reachable deformation space, we sweep across a dense grid of actuation commands to generate a large dataset of $\{(u_{\text{sim}}^{(i)}, Y_{\text{sim}}^{(i)})\}_{i=1}^{N_{\text{sim}}}$ pairs, where $u_{\text{sim}}^{(i)} \in \mathbb{R}^2$ is the servo command and $Y_{\text{sim}}^{(i)} \in \mathbb{R}^{548 \times 3}$ is the corresponding 548-vertex mesh exported directly from the simulator.

\begin{figure}[t]
  \centering
  \includegraphics[width=\columnwidth]{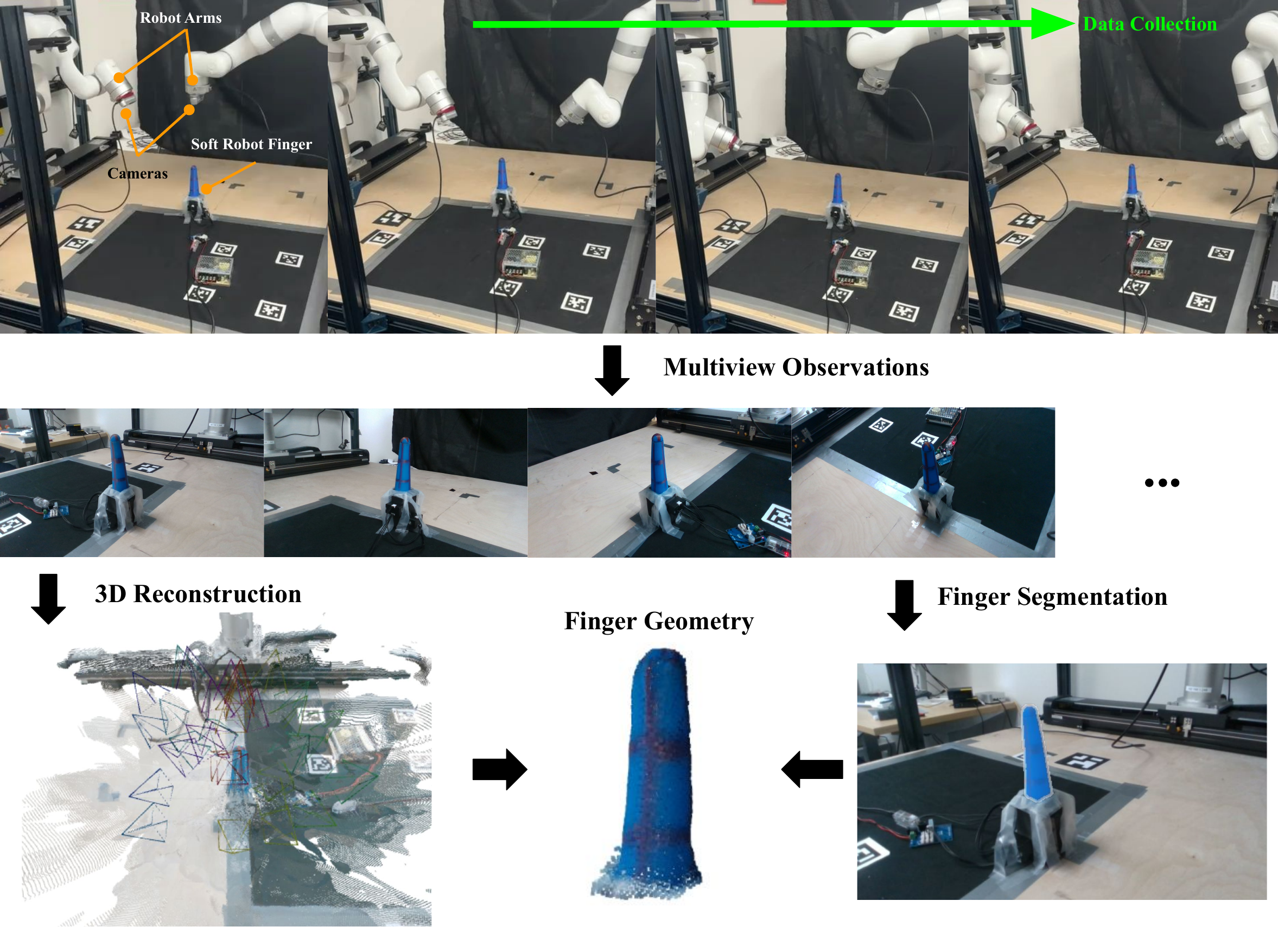}
  \caption{\textbf{Real-world data collection setup.} Two RGB cameras are mounted at distinct viewpoints around a soft finger actuated by an xArm 7 robot. For each servo command, synchronized multi-view images are captured and fed into our reconstruction pipeline to produce 3D point clouds used for real world deployment.}
  \label{fig:realworld_setup}
\end{figure}

\textbf{Real World.} We actuate the physical soft finger by two servo motors, each controlling one tendon pair, producing real servo commands $u_{\text{real}}^{(i)} \in \mathbb{R}^2$ measured in encoder ticks. To obtain 3D shape observations, we capture synchronized multi-view RGB images from two cameras mounted at fixed, calibrated viewpoints around the finger, as shown in Fig.~\ref{fig:realworld_setup}. By this setup, we ensure complete angular coverage of multiview captures for accurate 3D reconstruction while avoiding inter-view latency and the resulting multi-view inconsistencies.  We then apply VGGT~\cite{wang2025vggt} for multi-view reconstruction to preserve accurate geometric details of the finger, followed by SAM~\cite{ravi2024sam} for segmenting the finger from the reconstructed scene, to produce finger-only 3D point clouds $Y_{\text{real}}^{(i)}$ of the finger at each actuation state. Real-world data collection covers a representative grid of servo commands, yielding a dataset of paired observations $\{(u_{\text{real}}^{(i)}, Y_{\text{real}}^{(i)})\}_{i=1}^{N_{\text{real}}}$ for downstream alignment and residual training.

\subsection{Data Alignment}
To enable sim-to-real transfer, consistent pairing between simulation and real observations is essential~\cite{tian2025sim2real4deform}. However, since simulated and real shapes are produced by different pipelines, the raw observations differ in sampling density, vertex correspondence and scale, making directly matching ill-posed. To overcome this, we employ As-Rigid-As-Possible (ARAP)~\cite{sorkine2007rigid} alignment, which fits a common template mesh to observations from both simulated and real domains while preserving local rigidity, thus projecting the data into a shared, topologically-consistent state space. This yields stable correspondences that are robust to missing points and non-uniform sampling typical of real-world reconstructions.

Formally, let the template mesh be defined as $\mathcal{T}=(V_0,E)$, comprising rest vertices $V_0=\{v_i^0\}_{i=1}^{M}$ and edges $E$. Given an observed point cloud $P=\{p_k\}_{k=1}^{K}$, the ARAP registration solves for the deformed template vertices $V=\{v_i\}_{i=1}^{M}$ and optimal local rotations $\{R_i \in SO(3)\}$ by minimizing the following objective: %
\begin{equation}
\begin{aligned}
\min_{\{v_i\},\,\{R_i\}}\quad
&\sum_{(i,j)\in E} w_{ij}\,\big\lVert\,(v_i-v_j)-R_i\,(v_i^0-v_j^0)\,\big\rVert_2^2 \\
&\quad+\;\lambda \sum_{i=1}^{M}\big\lVert\,v_i-\Pi_{P}(v_i)\,\big\rVert_2^2 \;,
\end{aligned}
\label{eq:arap_align}
\end{equation}
where the first term enforces locally rigid deformations across the template edges (weighted by $w_{ij}$), and the second term anchors the deformed template to the observation via a nearest-neighbor projection $\Pi_{P}(\cdot)$ onto $P$, scaled by a regularization weight $\lambda$.

After fitting, we extract a fixed subset of $N\leq M$ handle vertices $\mathcal{H}\subset\{1,\dots,M\}$. Their corresponding 3D coordinates serve as our final aligned representation: 
\begin{equation}
\hat{Y}(P) \;=\; \{\,v_i\,\}_{i\in\mathcal{H}} \in \mathbb{R}^{N\times 3}.
\label{eq:arap_handles}
\end{equation}

\noindent This provides a compact, N-dimensional representation that maintains topological consistency across both simulation and reality.

Using the ARAP-aligned representation, we construct one-to-one correspondences between real and simulated samples by nearest-neighbor matching in shape space. For each real sample with actuation command $u_{\text{real}}^{(i)}$ and ARAP-aligned shape $\hat{Y}_{\text{real}}^{(i)}$, we search over a set of simulated candidates $\{(u_{\text{sim}}^{(j)}, \hat{Y}_{\text{sim}}^{(j)})\}$ and select the simulation input whose aligned shape is most similar under mean absolute error (MAE):
\[
j^{*}(i)=\arg\min_{j}\; \mathrm{MAE}\!\left(\hat{Y}^{(i)}_{\text{real}},\hat{Y}^{(j)}_{\text{sim}}\right),
\qquad
u^{*}_{\text{sim}}(i)=u^{(j^{*}(i))}_{\text{sim}}.
\]

\noindent We then fit a global calibration map that predicts the matched simulation command directly from the real servo command. Let $u_{\text{home}}$ denote the home tick position and define relative ticks $r^{(i)} = u_{\text{real}}^{(i)} - u_{\text{home}}$. We fit an affine mapping $g(r) = Ar + b$ to minimize the squared error between $g(r^{(i)})$ and the matched simulation commands $u_{\text{sim}}^*(i)$:
\[
u_{\text{sim}} \approx g(u_{\text{real}})=A\,(u_{\text{real}}-u_{\text{home}})+b.
\]

After this alignment procedure, every real sample is assigned a calibrated simulation-space command $u_{\text{sim}} \approx g(u_{\text{real}})$ and both domains share a consistent ARAP-aligned shape representation.

\subsection{Pre-training Physics Prior}
\label{sec:sim_pretrain}

Our goal is to learn a forward model that maps low-dimensional actuation to a geometrically accurate 3D finger shape. In practice, real-world reconstructions can exhibit noise and inconsistent sampling, which may obscure the underlying physics-consistent deformation behavior. We therefore pre-train a physics prior on clean simulation data before adapting the model to real observations.

Let $u \in \mathbb{R}^2$ denote the 2D actuation input (two coupled servo commands), and let $Y \in \mathbb{R}^{N\times 3}$ denote the corresponding finger surface geometry represented as a fixed-topology vertex set with $N=548$ vertices. We flatten the output as $y=\mathrm{vec}(Y)\in\mathbb{R}^{3N}$ and train a regression network
\begin{equation}
\hat{y} = f_{\theta}(u), \qquad f_{\theta}:\mathbb{R}^{2}\rightarrow\mathbb{R}^{3N},
\label{eq:forward_model}
\end{equation}
where $f_{\theta}$ is a multilayer perceptron (MLP) with parameters $\theta$ and $\hat{Y}=\mathrm{unvec}(\hat{y})\in\mathbb{R}^{N\times 3}$ is the predicted vertex set. The model architecture is a fully-connected MLP with five hidden layers of widths $[256,\,512,\,1024,\,512,\,256]$. Each hidden block applies a linear layer followed by Batch Normalization, a ReLU nonlinearity, and dropout ($p=0.1$) for regularization, while the final output layer is linear without activation to directly regress the $3N$ vertex coordinates. At inference time, the predicted vector $\hat{y}$ is reshaped back into $\hat{Y}\in\mathbb{R}^{N\times 3}$.

To stabilize optimization under the high-dimensional output, we normalize both inputs and outputs using dataset statistics computed over the simulation corpus. Specifically, with per-dimension mean and standard deviation $(\mu_u,\sigma_u)$ for $u$ and $(\mu_y,\sigma_y)$ for $y$, we train the network in normalized coordinates:
\begin{equation}
\tilde{u}=\frac{u-\mu_u}{\sigma_u+\epsilon},\qquad
\tilde{y}=\frac{y-\mu_y}{\sigma_y+\epsilon},\qquad
\hat{\tilde{y}} = f_{\theta}(\tilde{u}),
\label{eq:normalization}
\end{equation}
and recover predictions in metric space using
\begin{equation}
\hat{y} = \hat{\tilde{y}}\odot \sigma_y + \mu_y,
\label{eq:denorm}
\end{equation}
where $\odot$ denotes element-wise multiplication.

\begin{figure*}[t]
  \centering
  \includegraphics[width=\textwidth]{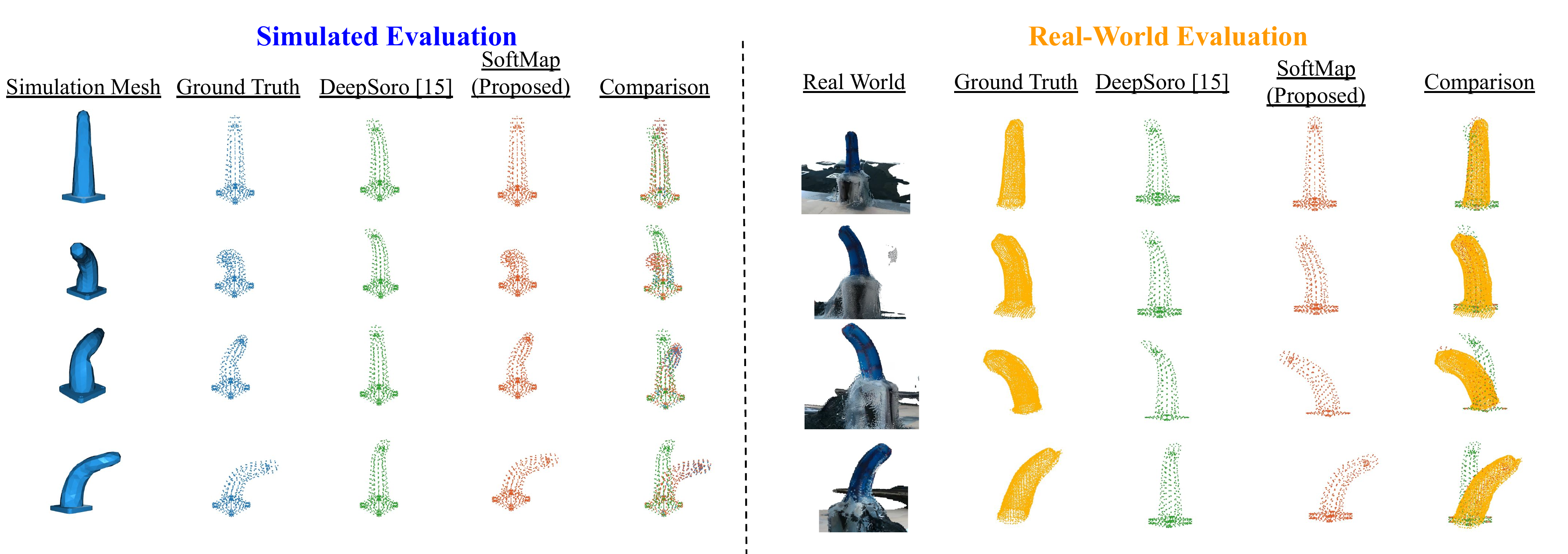}
  \caption{\textbf{Evaluation Comparison.} Left: Point Cloud Comparisons in Simulation Data; Right: Point Cloud Comparisons in Real Data}
  \label{fig:wide}
  \vspace{-2em}
\end{figure*}

We pre-train $f_{\theta}$ on simulation-generated deformations by minimizing a mean-squared error over vertices:
\begin{equation}
\mathcal{L}_{\text{sim}}(\theta)
=
\frac{1}{|\mathcal{D}_{\text{sim}}|}
\sum_{(u,y)\in\mathcal{D}_{\text{sim}}}
\left\|
f_{\theta}(\tilde{u})-\tilde{y}
\right\|_2^2.
\label{eq:sim_loss}
\end{equation}
Although the model does not explicitly integrate dynamics, it is physics-consistent in the sense that it is trained exclusively on simulation-consistent shapes produced by the underlying soft-body simulator; as a result, the learned mapping is smooth over the actuation space and outputs plausible deformations across the dataset range, which is critical for downstream inverse problems such as trajectory generation (Sec. \ref{sec:traj_gen}) and real-time teleoperation (Sec. \ref{sec:realtime_teleop}).

\begin{figure}[t]
  \centering
  \includegraphics[width=\columnwidth]{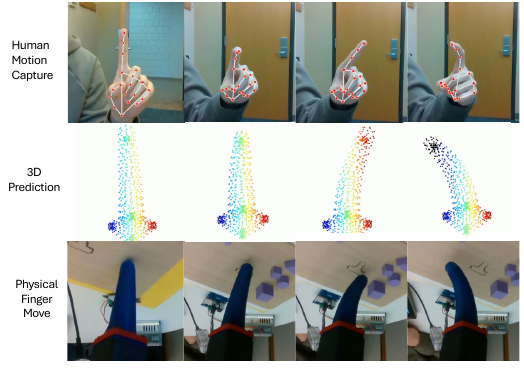}
  \caption{\textbf{Vision-based teleoperation demonstration.} Top: human motion capture with hand landmarks. Middle: real-time 3D shape prediction. Bottom: corresponding physical soft-finger motion.}
  \label{fig:teleop_demo}
\end{figure}
Vision-based teleoperation is increasingly common~\cite{qin2023anyteleop}, but applying it to soft robots remains challenging due to their highly compliant, hard-to-model deformations. Leveraging the learned forward shape predictor, we demonstrate a real-time teleoperation interface (Fig.~\ref{fig:teleop_demo}) that maps human motion to soft-finger actuation with continuous shape feedback.

\subsection{Residual Correction}
\label{sec:res_corr}
After pre-training the simulation model as a strong physics prior, we adapt it to real observations by learning a lightweight residual correction module. Given that the simulation model predicts a dense vertex set $V_{\text{sim}}=f_{\text{sim}}(u_{\text{sim}})\in\mathbb{R}^{N\times 3}$ with $N=548$, We then train a correction network to predict a per-vertex displacement field $\Delta_{\phi}(u_{\text{sim}})\in\mathbb{R}^{N\times 3}$, and form the real-domain prediction by
$V_{\text{real}}(u_{\text{sim}})=V_{\text{sim}}+\Delta_{\phi}(u_{\text{sim}}).$

This residual correction network is a compact MLP that maps the same 2D input to a $(548,3)$ displacement output, using LayerNorm, Gaussian Error Linear Unit (GELU) activations, and dropout for regularization. Here we replace BatchNorm and ReLU in Sec.~\ref{sec:sim_pretrain} with LayerNorm and GELU because the real-world residuals have small batch size and larger distribution shift. LayerNorm avoids dependence on batch statistics, and the smoother GELU nonlinearity better supports learning subtle, continuous displacement corrections. 

Because real observations are reconstructed as point clouds with variable sampling density and without vertex correspondence, we supervise the corrected prediction using the symmetric chamfer distance (CD) between the predicted vertices $V_{\text{real}}$ and the real point set $P_{\text{real}}$.

To prevent overfitting and to keep the refined shape close to the simulation prior, we regularize the magnitude of the displacement field,
\[
\mathcal{L}_{\text{l2}} = \frac{1}{N}\sum_{i=1}^{N}\|\Delta_i\|_2^2,
\]
and encourage spatial smoothness by penalizing differences across mesh edges $E$ (from a reference template mesh),
\[
\mathcal{L}_{\text{lap}} = \frac{1}{|E|}\sum_{(i,j)\in E}\|\Delta_i-\Delta_j\|_2^2.
\]
The final training objective is a weighted combination of the chamfer loss and the regularizers:
\[
\mathcal{L}=\mathrm{CD}\!\left(V_{\text{real}},P_{\text{real}}\right)+\lambda_{\text{l2}}\mathcal{L}_{\text{l2}}+\lambda_{\text{lap}}\mathcal{L}_{\text{lap}}.
\]
This residual formulation yields an accurate real-domain forward model while retaining the simulation-trained physics prior, providing a reliable foundation for downstream control tasks such as trajectory generation and teleoperation.

\subsection{Trajectory Tracking}
\label{sec:traj_gen}

Real-time trajectory tracking is a fundamental task to evaluate the capability for soft robots' control. 
Building on the calibrated forward model introduced above, we use the network as a fast differentiable predictor to enable model-based trajectory generation. Rather than manually tuning servo commands, we specify a desired fingertip path in Cartesian space and solve for the corresponding sequence of 2D actuation commands that reproduces this motion in both simulation and on hardware.

To define a consistent task-space reference, we reduce each predicted shape to a single fingertip point. We identify the fingertip $\mathbf{p}_{\text{tip}}(u)\in\mathbb{R}^3$ as the centroid of the vertices that experience the largest deformation under actuation. This fingertip point is tracked over time to form the executed trajectory.

We generate target trajectories as parametric curves centered at the rest fingertip position, parameterized by amplitude (mm), duration (s), and waypoint rate (Hz), producing a time-ordered set of waypoints $\{\mathbf{p}^*_t\}_{t=1}^{T}$. For each waypoint $\mathbf{p}^*_t$, we solve a 2D inverse-kinematics problem using the learned forward model by selecting an actuation command $u_t$ within bounds $[\boldsymbol{\ell},\mathbf{h}]$ that minimizes the distance between the model-predicted fingertip and the target. When a trajectory is defined in a lower-dimensional plane, we apply a projection $\Pi$ onto the active axes:
\[
u_t=\arg\min_{u\in[\boldsymbol{\ell},\mathbf{h}]}\left\lVert \Pi\!\left(\mathbf{p}_{\text{tip}}(u)-\mathbf{p}^*_t\right)\right\rVert_2.
\]
Because this objective is generally non-convex, we adopt a hybrid solver. For the first waypoint, we perform a coarse grid search over the actuation bounds to obtain a robust initialization, followed by Nelder–Mead refinement~\cite{Nelder1965ASM} and a short gradient-based polish using the Adam optimizer by differentiating through the network. For subsequent waypoints, we skip the grid search by warm-starting from the previous solution and refine each $u_t$ using local optimization, which yields temporally smooth command sequences with low per-waypoint computation. The resulting actuation sequence $\{u_t\}$ can then be replayed in simulation and on hardware.

\begin{table}[t]
\centering
\caption{Shape prediction evaluation in simulation and real settings (lower is better). Mean vertex error is not reported for real data as point cloud observations lack vertex-level correspondence.}
\label{tab:shape}
\small
\setlength{\tabcolsep}{2.2pt}
\begin{tabular}{l c c}
\toprule
Model & {Chamfer $\downarrow$ (mm)} & {Mean vertex $\downarrow$ (mm)} \\
\midrule
\multicolumn{3}{l}{\textit{Simulation}} \\
Laplacian-Encoded Model & $8.739\pm3.67$ & $8.757\pm4.811$ \\
DeepSoRo                & $4.520\pm2.437$ & $7.970\pm4.127$ \\
Linear Model            & $2.726\pm1.043$ & $2.620\pm1.143$ \\
XGBoost Model           & $1.28\pm0.58$ & $0.868\pm0.563$ \\
ARAP-Encoded Model      & $0.867\pm0.53$ & $0.438\pm0.274$ \\
\textbf{\algname (Proposed)} & $\mathbf{0.389\pm0.188}$ & $\mathbf{0.196\pm0.098}$ \\
\midrule
\multicolumn{3}{l}{\textit{Real}} \\
DeepSoRo                & $6.386\pm1.367$ & -- \\
\algname (w/o residual) & $5.681\pm0.939$ & -- \\
\textbf{\algname (Proposed)} & $\mathbf{3.786\pm0.61}$ & -- \\
\bottomrule
\end{tabular}
\end{table}

\subsection{Real-Time Visual Teleoperation}
\label{sec:realtime_teleop}

\begin{figure*}[t]
  \centering
  \includegraphics[width=0.9\textwidth]{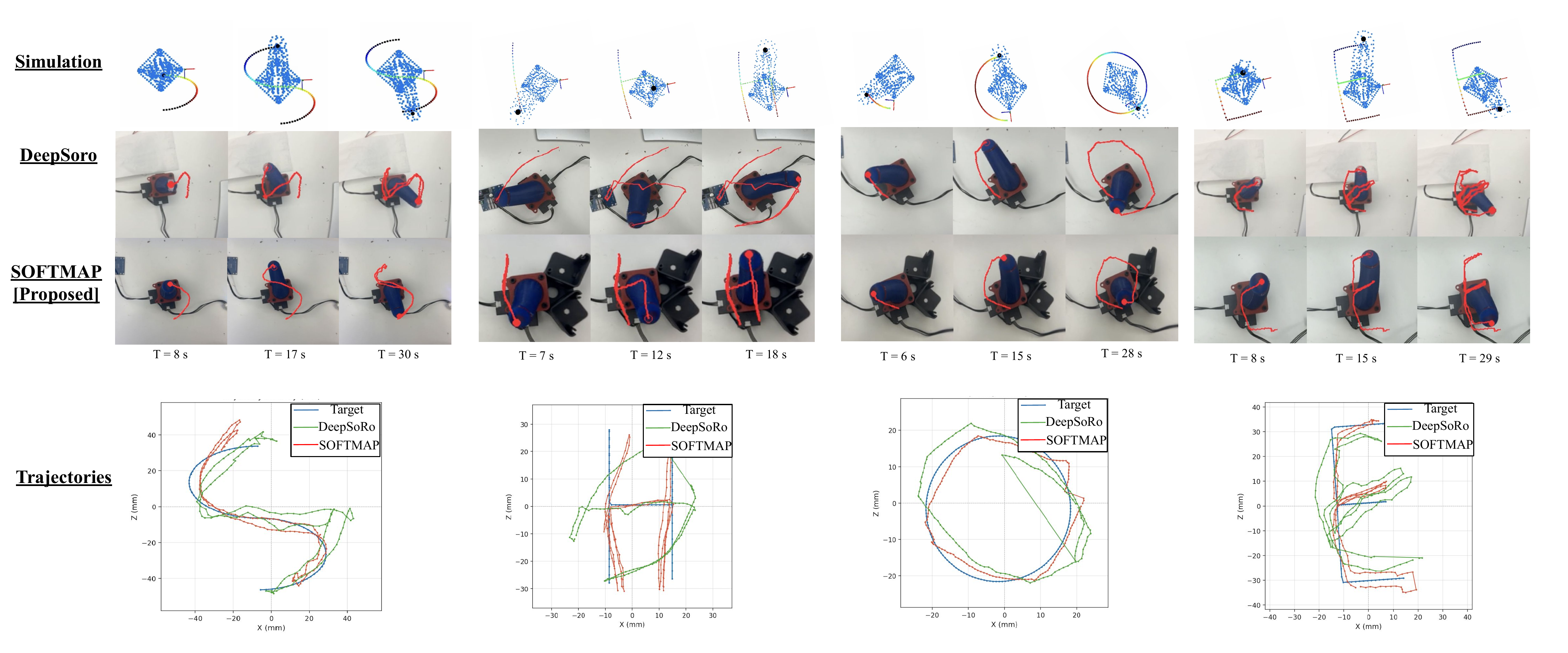}
  \caption{\textbf{Qualitative trajectory tracking results.} Each column shows snapshots at three timesteps for trajectories S, H, O, and E, comparing \algname against the DeepSoRo baseline and simulation ground truth. \algname consistently produces tighter fingertip paths with less drift and distortion, particularly on asymmetric strokes.} 
  \label{fig:trajQual}
  \vspace{-1em}
\end{figure*}


  


\begin{table*}[t]
\centering
\caption{Trajectory tracking error comparison (lower is better). Each trajectory is evaluated in simulation (Sim) and on real hardware (Real) using Mean Squared Error (MSE) $\pm$ standard deviation and maximum error, all in mm.}
\small
\setlength{\tabcolsep}{6pt}
\renewcommand{\arraystretch}{1}
\begin{tabular}{l l cc cc}
\toprule
\multirow{2}{*}{Trajectory} &
\multirow{2}{*}{Method} &
\multicolumn{2}{c}{Sim} &
\multicolumn{2}{c}{Real} \\
\cmidrule(lr){3-4}\cmidrule(lr){5-6}
& & MSE [mm]  & Max [mm] & MSE [mm]  & Max [mm] \\
\midrule
\multirow{2}{*}{Letter S}
& DeepSoRo        & $2.685\pm1.109$   & 3.646  & $5.917\pm5.12$  & 23.154 \\
& \algname (Proposed) & $\mathbf{1.124\pm0.617}$ & $\mathbf{1.95}$  & $\mathbf{3.458\pm2.976}$ & $\mathbf{15.38}$ \\
\midrule
\multirow{2}{*}{Letter H}
& DeepSoRo        & $2.766\pm0.999$   & 4.319  & $5.854\pm3.816$ & 14.7 \\
& \algname (Proposed) & $\mathbf{1.214\pm0.652}$  & $\mathbf{2.55}$ & $\mathbf{3.569\pm2.343}$ & $\mathbf{10.3}$ \\
\midrule
\multirow{2}{*}{Letter O}
& DeepSoRo        & $1.336\pm0.918$   & 1.582  & $3.164\pm1.683$ & 6.887 \\
& \algname (Proposed) & $\mathbf{1.184\pm0.182}$ & $\mathbf{1.499}$ & $\mathbf{1.387\pm0.831}$ & $\mathbf{3.839}$ \\
\midrule
\multirow{2}{*}{Letter E}
& DeepSoRo        & $1.889\pm0.92$    & 3.015  & $4.791\pm3.174$ & 14.76 \\
& \algname (Proposed) & $\mathbf{1.174\pm0.755}$ & $\mathbf{2.509}$  & $\mathbf{2.616\pm1.777}$ & $\mathbf{8.05}$ \\
\bottomrule
\end{tabular}
\label{tab:traj_compare}
\vspace{-1.5em}
\end{table*}

\subsubsection{Human Motion Capture}
We capture human finger motion using an Intel RealSense RGB stream (640×480 at 30 FPS) and estimate hand pose using MediaPipe Hands. For each frame, MediaPipe provides 21 3D landmarks per detected hand. We use the wrist and middle Metacarpophalangeal (MCP) landmarks $\mathbf{p}_{wrist}, \mathbf{p}_{mid}$ to define a stable forward direction of the palm in the image plane, and the index MCP and index fingertip landmarks $\mathbf{p}_{mcp}, \mathbf{p}_{tip}$ to measure index-finger motion relative to the palm. These landmarks define the basic displacement signal $\mathbf{d}=\mathbf{p}_{\mathrm{tip}}-\mathbf{p}_{\mathrm{mcp}}$.

\subsubsection{From Human Motion to Model Actuation}
To convert tracked landmarks into model actuation, we compute a hand-local coordinate frame in the image plane to obtain control signals that are robust to hand orientation and distance from the camera. We define the forward unit vector $\hat{\mathbf{f}}$ using the wrist-to-middle-MCP direction and the right unit vector $\hat{\mathbf{r}}$ as its in-plane perpendicular:
\begin{equation}
\hat{\mathbf{f}}=
\frac{\mathbf{p}_{\mathrm{mid}}-\mathbf{p}_{\mathrm{wrist}}}{\|\mathbf{p}_{\mathrm{mid}}-\mathbf{p}_{\mathrm{wrist}}\|_2+\epsilon},
\qquad
\hat{\mathbf{r}}=
\begin{bmatrix}
\hat{f}_y\\
-\hat{f}_x
\end{bmatrix}.
\label{eq:hand_frame}
\end{equation}
Using the fingertip displacement $\mathbf{d}$, we compute scale-normalized projections $c_f$ (forward component) and $c_r$ (right component) by dividing the hand scale $s$ to remove scale changes caused by the user moving closer to or farther from the camera.
\begin{equation}
s=\|\mathbf{p}_{\mathrm{mid}}-\mathbf{p}_{\mathrm{wrist}}\|_2+\epsilon,
\qquad
c_f=\frac{\mathbf{d}^\top \hat{\mathbf{f}}}{s},
\qquad
c_r=\frac{\mathbf{d}^\top \hat{\mathbf{r}}}{s}.
\label{eq:proj}
\end{equation}
We then map these components to the actuation commands $(u_{lat}, u_{vert})$ by scaling with the actuation range. This produces direct and continuous actuation commands in the same units used by our forward model.

\subsubsection{Real-Time Shape Prediction and Control}
At each timestep, we query \algname's model to predict the full 3D finger geometry. This predicted shape serves as a real-time proprioceptive state estimate for the soft finger, enabling accurate fingertip localization. To actuate the physical finger, we convert the model-space commands into servo goal positions. We compute goal ticks using the linear calibration layer that maps millimeter actuation to tick displacement. Each command is then hard-clamped to remain within a fixed safety window by setting the limit of ticks around the home position before being transmitted to the servos. This safety clamp prevents aggressive teleoperation gestures from exceeding mechanical limits or stressing the soft structure.

\section{Evaluation}
We evaluate \algname across three tasks of increasing complexity. First, we assess 3D shape prediction accuracy in both simulated and real settings, including an ablation of the residual correction network. Second, we evaluate the learned forward model on continuous fingertip trajectory tracking in both simulation and on hardware. Finally, we demonstrate the practical utility of the full system through real-time vision-based teleoperation on contact-rich pushing tasks.



\subsection{Shape Prediction}

\label{sec:shape_eval}

We evaluate shape prediction performance in both simulation and real settings. In all experiments, the model predicts a fixed-topology vertex set with $N=548$ vertices in $\mathbb{R}^3$, and we report (i) symmetric Chamfer distance (mm) between predicted vertices and the target point set, and (ii) mean per-vertex error (mm) in the simulation environment. 

\subsubsection{Simulation evaluation.}
We first evaluate the forward model trained on simulation data only. As shown in Table~\ref{tab:shape}, \algname achieves the lowest error among all baselines, with Chamfer distance of $0.389$\,mm and mean vertex error of $0.196$\,mm. We compare against alternative encodings and regressors trained on the same dataset, including ARAP-encoded and Laplacian-encoded~\cite{sorkine2004laplacian} representations as well as linear and XGBoost models. While ARAP encoding remains competitive, Laplacian encoding and simpler regressors substantially degrade accuracy, indicating that preserving the full geometry in a dense vertex representation is important for precise shape prediction under actuation. We additionally compare against DeepSoRo~\cite{yoo2023toward}, which performs worse in this setting, suggesting that the forward prediction model is well-suited for the simulator’s deformation distribution.

\subsubsection{Real-World Evaluation}
Next, we evaluate sim-to-real residual correction by comparing predictions to real reconstructed point clouds. As shown in Table~\ref{tab:shape}, sim-only models exhibit a clear sim-to-real gap, which we attribute to unmodeled physical nonlinearities such as material hysteresis and manufacturing variability that are not captured by the SOFA simulation. Applying residual learning (Sec.~\ref{sec:res_corr}) reduces this gap by 33.4\%, relative to the sim-only \algname model, and outperforms DeepSoRo by 40.7\% in Chamfer distance on real data, despite being trained on a significantly smaller real-world dataset. Notably, even the sim-only \algname model without residual correction outperforms DeepSoRo, suggesting that the ARAP-aligned physics prior alone provides a strong geometric foundation. The additional residual correction then refines this foundation by learning systematic per-vertex corrections that account for the remaining domain shift. These results indicate that (i) the simulation-trained \algname model provides a strong geometric prior, (ii) a lightweight residual correction network is an effective and data-efficient mechanism for bridging the remaining sim-to-real gap, and (iii) the two components are complementary, each contributing to the accuracy.

In Fig.~\ref{fig:wide}, we provide qualitative comparisons of predicted shapes against ground-truth point clouds in both simulation and real settings. Across a range of rest and bent configurations, \algname produces point clouds that closely match the overall curvature and global pose of the finger, with noticeably smaller geometric deviation than baseline. The overlap visualizations further highlight that our predictions maintain better spatial alignment with the ground truth, especially in highly curved states where small actuation errors can lead to large tip displacement. DeepSoRo, by contrast, struggles particularly in extreme bend configurations, where its predictions diverge significantly from the groundtruth geometry. Consistent with the quantitative results, these qualitative comparisons indicate that \algname better preserves the underlying geometry across domains, and that the combination of ARAP alignment and residual correction is key to achieving accurate real-world shape prediction.

\subsection{Trajectory Tracking}
\label{sec:traj_eval}

We evaluate trajectory tracking by commanding the soft finger to follow four target fingertip paths (letters S/H/O/E) and measuring the Euclidean tracking error between the desired waypoint $\mathbf{p}^*_t$ and the executed fingertip position $\mathbf{p}_t$ in millimeters. Each trajectory is tested both in simulation and on hardware using the same solved command sequence.

Across all trajectories, \algname consistently outperforms DeepSoRo in both simulation and real-hardware tracking (Table~\ref{tab:traj_compare}). On real hardware, \algname achieves substantially lower mean errors and reduces worst-case deviations across all letter trajectories, which matches the qualitative overlays in Fig.~\ref{fig:trajQual} where DeepSoRo exhibits larger drift and distortion on asymmetric strokes. In simulation, \algname also shows markedly smaller residual errors, indicating a tighter coupling between the learned forward model and our inverse-kinematics solver. This improved model–solver consistency carries over to more reliable execution on the physical system.

\subsection{Teleoperation}

\begin{figure}[t]
  \centering
  \includegraphics[width=\columnwidth]{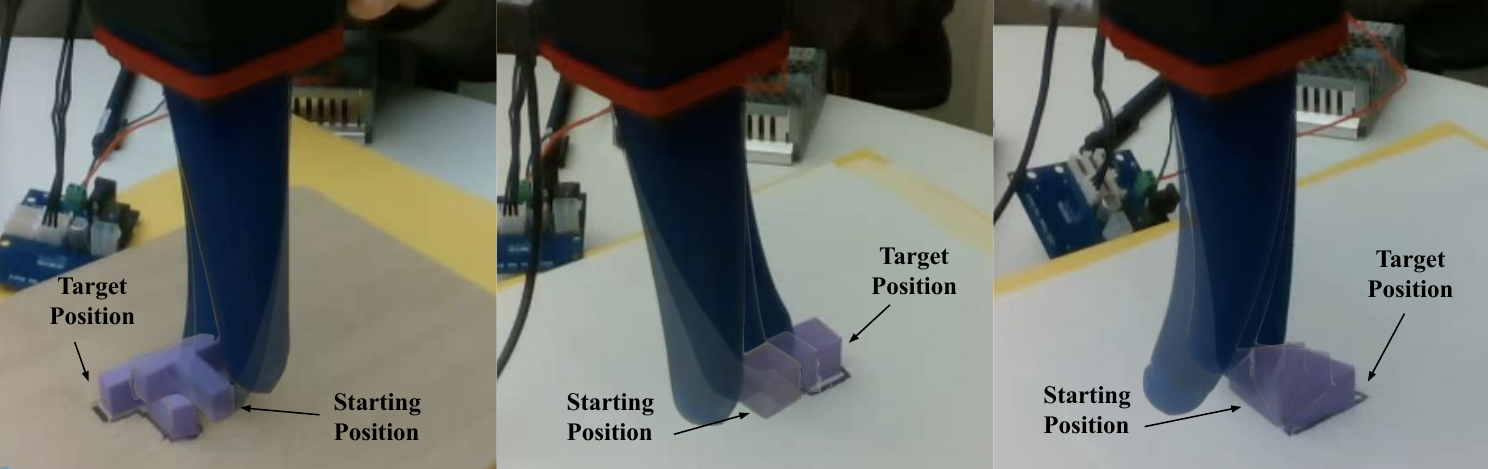}
  \caption{\textbf{Teleoperation task progression.} From left to right, the soft finger pushes the T, Cube and a triangle prism from its starting pose toward the marked target region.}
  \label{fig:push-tasks}
\end{figure}

Inspired by robot control benchmarks~\cite{chi2025diffusion}, we evaluate vision-based teleoperation on three pushing tasks (letter T, cube, and triangular prism), where the goal is to push a target object into a goal region from different directions (Fig.~\ref{fig:push-tasks}). For each trial, we compute (i) the intersection-over-union (IoU) between the final object footprint and the goal region and (ii) an overlap percentage measuring how much of the object lies inside the goal region (higher is better). As summarized in Table~\ref{tab:task_iou_overlap_grouped}, \algname consistently outperforms DeepSoRo across all tasks. The largest gains are observed on Push-T, the most geometrically constrained task, where \algname improves IoU by 89.5\% and overlap by 73.9\%, suggesting that accurate 3D shape prediction is particularly critical when precise fingertip placement is required. More symmetric objects such as the cube and triangular prism show smaller but consistent improvements of 27.0\% and 10.9\% in IoU respectively, indicating that \algname's shape-aware forward model generalizes across object geometries. Overall, averaged across tasks, \algname improves IoU by 36.5\% and overlap by 29.9\% relative to DeepSoRo, indicating more accurate control in real-time teleoperation.

The evaluation indicates that improving 3D shape prediction directly benefits downstream manipulation. \algname provides real-time estimates of the full finger geometry, enabling more accurate fingertip localization and reducing off-axis contacts and drift during contact-rich pushing. Overall, these results suggest that shape-aware sim-to-real modeling yields more reliable control for soft manipulation.

\begin{table}[t]
\centering
\caption{Teleoperation push-task comparisons (higher is better).}
\label{tab:task_iou_overlap_grouped}
\scriptsize
\setlength{\tabcolsep}{3pt}      
\renewcommand{\arraystretch}{1.08}

\begin{tabular}{p{0.30\columnwidth} l c c}
\toprule
\textbf{Task} & \textbf{Method} & \textbf{IoU $\uparrow$} & \textbf{Overlap $\uparrow$ (\%)} \\
\midrule
\multirow{2}{*}{Push-T}
  & DeepSoRo & $0.38\pm0.26$ & $46\pm27$ \\
  & \algname\ (Proposed) & $\mathbf{0.72\pm0.13}$ & $\mathbf{80\pm11}$ \\
\midrule
\multirow{2}{*}{Cube}
  & DeepSoRo & $0.63\pm0.16$ & $71\pm15$ \\
  & \algname\ (Proposed) & $\mathbf{0.80\pm0.10}$ & $\mathbf{87\pm8}$ \\
\midrule
\multirow{2}{*}{Triangular Prism}
  & DeepSoRo & $0.55\pm0.10$ & $67\pm6.5$ \\
  & \algname\ (Proposed) & $\mathbf{0.61\pm0.18}$ & $\mathbf{72\pm15}$ \\
\bottomrule
\end{tabular}
\end{table}

\section{Conclusion and Future Work}
We present \algname, a sim-to-real learning framework for accurate, real-time 3D forward modeling of tendon-actuated soft finger manipulators. By combining ARAP-based topological alignment, simulation pretraining, lightweight residual correction, and a closed-form actuation calibration layer, \algname bridges the reality gap with minimal real-world data. Experiments demonstrate state-of-the-art shape prediction accuracy in both simulation and real settings, fingertip trajectory tracking across multiple target paths, and improved teleoperation performance.

Despite these results, several limitations remain. The current framework assumes a fixed finger morphology and requires re-collection of real-world data and retraining of the residual correction network when the hardware changes. Additionally, while the forward model enables accurate fingertip tracking, it does not yet account for contact forces, which are critical for robust manipulation. Future work will explore extending \algname to multi-finger systems and dexterous hands, where inter-finger interactions introduce additional modeling complexity. Incorporating contact-aware modeling and force feedback could further improve reliability in contact-rich tasks. More expressive network architectures or uncertainty-aware models could also improve generalization under larger sim-to-real gaps or across different soft designs.

\bibliographystyle{IEEEtran}
\bibliography{references}

\end{document}